\documentclass[conference]{IEEEtran}
\ifCLASSINFOpdf
\else
\fi
\usepackage{times}
\usepackage{epsfig}
\usepackage{graphicx}
\usepackage{amsmath}
\usepackage{amssymb}
\usepackage{booktabs}
\usepackage{color}
\usepackage{colortbl}
\usepackage{subfig}
\usepackage{flushend}

\definecolor{cwblue1}{rgb}{0.27,0.427,0.623}
\definecolor{cwblue2}{rgb}{0.286,0.454,0.658}
\definecolor{cwblue3}{rgb}{0.733,0.811,0.905}

\def\NA{} 

\begin{document}
%
\title{Full-Page Text Recognition: \\ Learning Where to Start and When to Stop}



\author{
\IEEEauthorblockN{Bastien Moysset\IEEEauthorrefmark{1}\IEEEauthorrefmark{4}, Christopher Kermorvant\IEEEauthorrefmark{2}, Christian Wolf\IEEEauthorrefmark{3}\IEEEauthorrefmark{4}}
\IEEEauthorblockA{\IEEEauthorrefmark{1}A2iA SA, Paris,  France}
\IEEEauthorblockA{\IEEEauthorrefmark{2}Teklia SAS, Paris,  France}
\IEEEauthorblockA{\IEEEauthorrefmark{3}Universit\'e de Lyon, CNRS, France }
\IEEEauthorblockA{\IEEEauthorrefmark{4}INSA-Lyon, LIRIS, UMR5205, F-69621}
%
%
}


%


\maketitle

\begin{abstract}

Text line detection and localization is a crucial step for full page document analysis, but still suffers from heterogeneity of real life documents. In this paper, we present a new approach for full page text recognition. Localization of the text lines is based on regressions with Fully Convolutional Neural Networks and Multidimensional Long Short-Term Memory as contextual layers.

In order to increase the efficiency of this localization method, only the position of the left side of the text lines are predicted. The text recognizer is then in charge of predicting the end of the text to recognize. This method has shown good results for full page text recognition on the highly heterogeneous Maurdor dataset.
\end{abstract}


%
\IEEEpeerreviewmaketitle

\section{Introduction}

Most applications in document analysis require text recognition at page level, where only the raw image is available and no preliminary hand-made annotation can be used. Traditionally, this problem has mainly been addressed by separating the process into two distinct steps; namely the text line detection task, which is frequently proceeded by additional paragraph and word detection steps, and the text recognition task. In this work we propose a method, which couples these two steps tighter by unloading some of the burden of the difficult localization step to the recognition task. In particular, the localization step detects the starts of the text lines only. The problem of finding where to stop the recognition is solved by the recognizer itself.

\subsection{Related work}
Numerous algorithms have been proposed for the text line localization. Some are used in a bottom-up approach by grouping sub-components like connected components or black pixels into lines. RLSA \cite{wong1982document} uses morphological opening on black pixels to merge the components that belong to the same text line. Similarly, Shi et al. \cite{Shi2009a} resort to horizontal ellipsoidal steerable filters to blur the image and merge the components of the text line. In \cite{WolfICPR2002V}, gradients are accumulated and filtered. Louloudis et al. \cite{Louloudis2009b} employ a Hough algorithm on the connected component centers while Ryu et al. \cite{ryu2014language} cluster parts of the connected components according to heuristic based successive splits and merges.

Other methods follow a top-down approach and split the pages into smaller parts. The XY-cut algorithm \cite{nagy1984hierarchical} looks for vertical and horizontal white spaces to successively split the pages in paragraphs, lines and words. Similarly, projection profile algorithms like Ouwayed et al. \cite{ouwayed2012general} are aimed at finding the horizontal whiter parts of a paragraph. This technique is extended to non-horizontal texts by methods like Nicolaou et al. \cite{Nicolaou2009} that dynamically finds a path between the text lines or by Tseng et al. \cite{tseng1999recognition} that use a Viterbi algorithm to minimize this path.

Techniques like the ones proposed by Mehri et al. \cite{mehri2013pixel} or Chen et al. \cite{chen2015page} classify pixels into text or non-text but need post-processing techniques to constitute text lines.

These techniques usually work well on the homogeneous datasets they have been tuned for but need heavy engineering to perform well on heterogeneous datasets like the Maurdor dataset \cite{Brunessaux2014}. For this reason, Machine learning has proven to be efficient, in particular deep convolutional networks. Early work from Delakis et al. \cite{delakis2008text} classifies scene text image parts as text and non-text with a Convolutional Neural Network on a sliding window. In \cite{moysset2015paragraph}, paragraph images are split vertically using a recurrent neural network and CTC alignment. More recently, methods inspired from image object detection techniques like MultiBox \cite{erhan2014scalable}, YOLO \cite{YOLOCVPR2016} or Single-Shot Detector (SSD) \cite{LiuErhanSSD2016} have arisen. Moysset et al. \cite{moysset2016learning} proposed a MultiBox based approach for direct text line bounding boxes detection. Similarly, Gupta et al. \cite{gupta2016synthetic} and Liao et al. \cite{liao2016textboxes} use respectively YOLO based and SSD based approach for scene text detection. Moysset et al. \cite{moysset2016points} also propose the separate detection of bottom-left and top-right corners of line bounding boxes.

The text recognition part is usually made with variations of Hidden Markov Models \cite{bertolami2008hidden} or 2D Long Short Term Memory (2D-LSTM) \cite{Graves2DLSTM2009} neural networks.

Finally, Bluche et al. \cite{BlucheNIPS2016} use a hard attention mechanism to directly perform full page text recognition without prior localization. The iterative algorithm finds the next attention point based on the sequence of seen glimpses modeled through the hidden state of a recurrent network.

\subsection{Method overview}
In this work, we address full page text recognition in two steps. First, a neural network detects where to start to recognize a text line, and a second network performs the text recognition and decides when to stop the process. More precisely, the former network detects the left sides of each text lines by predicting the value of the object position coordinates as a regression problem. This detection neural network system is detailed in Part \ref{sec:objectLocalization} and the left-side strategy is explained in Part \ref{sec:triplets}. The latter network recognizes the text and predicts the end of the text of the line  as described in Part \ref{sec:recognizer}. The experimental setup is described in Part \ref{sec:experimental} and results are shown and analyzed in Part \ref{sec:results}.

\section{Object localization with deep networks}
\label{sec:objectLocalization}

\subsection{Network description}

\begin{table}
\begin{center}
\caption{Network architecture/hyper-parameters. The input and feature map sizes are an illustrative example. The number of parametres does \emph{NOT} depend on the size of the input image.}
\label{tab:architecture}
\begin{tabular}{lcccr}
Layer & Filter  & Stride & Size of the  & Number of  \\ 
      & size    &        & feature maps & parameters \\
\arrayrulecolor{cwblue1} \toprule  
Input & / & / & 1$\times$(598$\times$838) & \NA \\
C1  & 4$\times$4 & 3$\times$3 & 12$\times$(199$\times$279) & 204 \\
LSTM1 & / & / & " " & 8880 \\
C2 & 4$\times$3 & 3$\times$2 & 16$\times$(66$\times$139) & 2320 \\
LSTM2 & / & / & " " & 15680 \\
C3 & 6$\times$3 & 4$\times$2 & 24$\times$(16$\times$69) & 6936 \\
LSTM3 & / & / & " " & 35040 \\
C4 & 4$\times$3 & 3$\times$2 & 30$\times$(5$\times$34) & 8670 \\
LSTM4 & / & / & " " & 54600 \\
C5 & 3$\times$2 & 2$\times$1 & 36$\times$(2$\times$33) & 6516 \\
Output & 1$\times$1 & 1$\times$1 & 4$\times$20$\times$(2$\times$33) & 2960 \\
\end{tabular}
\end{center}
\end{table}

In the lines of \cite{erhan2014scalable}, we employ a neural network as a regressor to predict the positions of objects in images. The network predicts a given number $N$ of object candidates. Each of these object candidates is indexed by a linear index $n$ and defined by $K$ coordinates $l_n{=}\{ l^{k}_n \}, \ k{=}{1\dots4K}$ corresponding to the position of the object in the document and a confidence score $c_n$. As the number of objects in an image is variable, at test time, only the objects with a confidence score over a threshold are kept.

In order to cope with the small amount of training data available for   document analysis tasks and to detect a large number of objects corresponding to our text lines, we adopted the method described in \cite{moysset2016learning}. We do not use a fully connected layer at the end of the network that has as inputs features conveying information about the whole page image and, as outputs, all the object candidates of the page. Instead, our method is fully convolutional, which allows the network to share parameters over the different regressors. More precisely, we use a $1{\times}1$ convolution to predict the objects locally and, consequently, to highly reduce the number of parameters in the network. 

Layers constituted of Two-Dimensional Long-Short-Term-Memory cells (2D-LSTM) \cite{Graves2DLSTM2009} are interleaved between the convolutional layers in order to recover the context information lost by the local nature of the detection. 

\begin{figure}[!t]
  \centering
  \includegraphics[width=\linewidth]{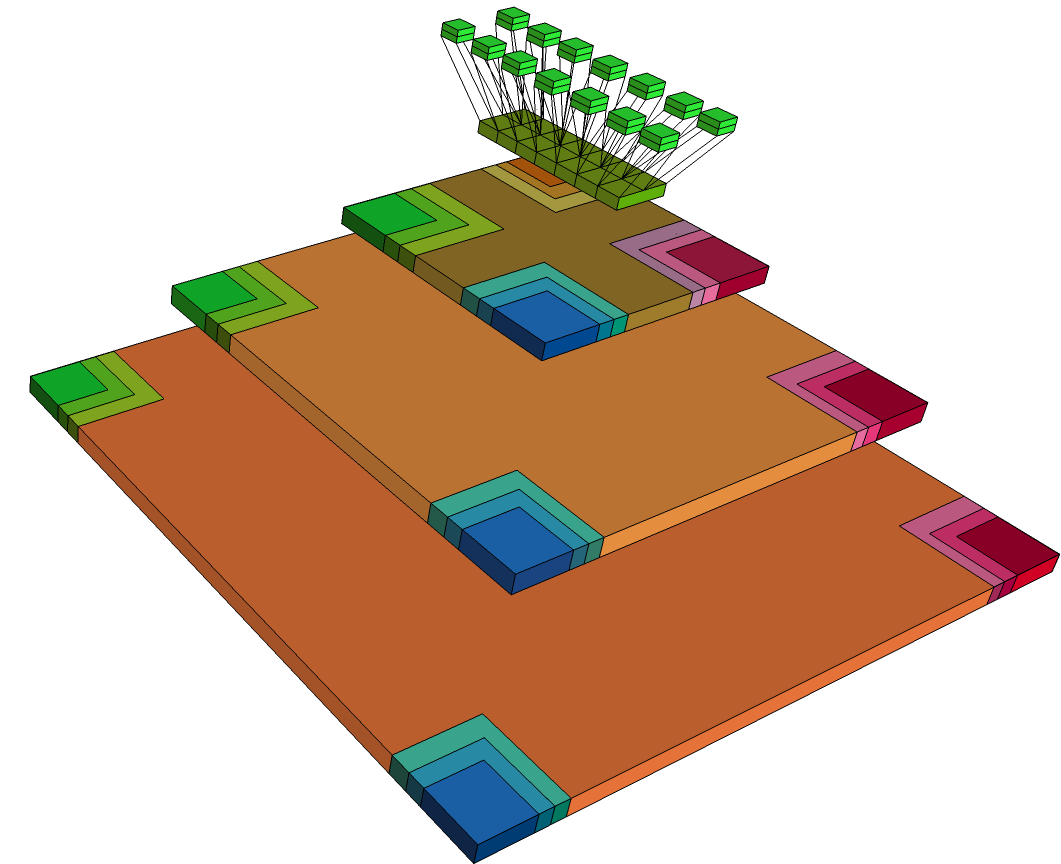}
  \caption{Sketch of the Convolutional Recurrent Neural Network that locally predicts the object positions (we do not show the correct numbers of layers and units).}
  \label{fig:archi}
\end{figure}

The architecture is similar to the one in \cite{moysset2016learning}. It is described in Table \ref{tab:architecture} and illustrated in Figure \ref{fig:archi}.

\subsection{Training}
We used the same training process as the one described in \cite{erhan2014scalable}. The cost function is a weighted sum between a confidence cost and the Euclidean distance between the two object positions (predicted and ground-truth):
\begin{equation}
\label{eq:globalCost}
\begin{split}
Cost & = \sum_{n=0}^N \sum_{m=0}^M X_{nm} \left( \alpha \left\|l_{n}-t_{m}\right\|^2 - \log(c_{n})\right) \\
 & - (1 - X_{nm}) \log(1-c_{n})
\end{split}
\end{equation}
Here, the $N$ object candidates have position coordinates $l_{n}$ and confidence $c_{n}$ while the $M$ reference objects have position coordinates $t_{m}$. $\alpha$ is a parameter weighting localisation and confidence costs.  As the output of the network (as well as the ground-truth information) is structured, a matching between the two of them is necessary in order to calculate the loss in equation \ref{eq:globalCost}. This matching is modelled through variable $X{=}\{X_{nm}\}$, a binary matrix. In particular, $X_{nm}{=}1$ if network output $n$ has been matched to ground truth object $m$ in the given image. Equation \ref{eq:globalCost} needs to be minimized under constraints enforcing one-to-one matches, which can be solved efficiently through the Hungarian algorithm \cite{munkres1957algorithms}.

We could not confirm the claims reported in \cite{erhan2014scalable} who apply this matching process for object detection in natural images. In particular, no improvement was found when using anchor positions associated to  objects which were mined through k-means clustering. On the other hand, we found it useful to employ different weights $\alpha$ for the two different uses of equation \ref{eq:globalCost}. A higher value for $\alpha$ was used during matching (solving for $X$) than for backpropagation (learning of network parameters). This favours the use of all outputs during training --- details are given in section \ref{sec:results}.

\section{Localization and recognition}

\begin{figure}[!t]
  \centering
  \small{(a)} \includegraphics[width=2.5in]{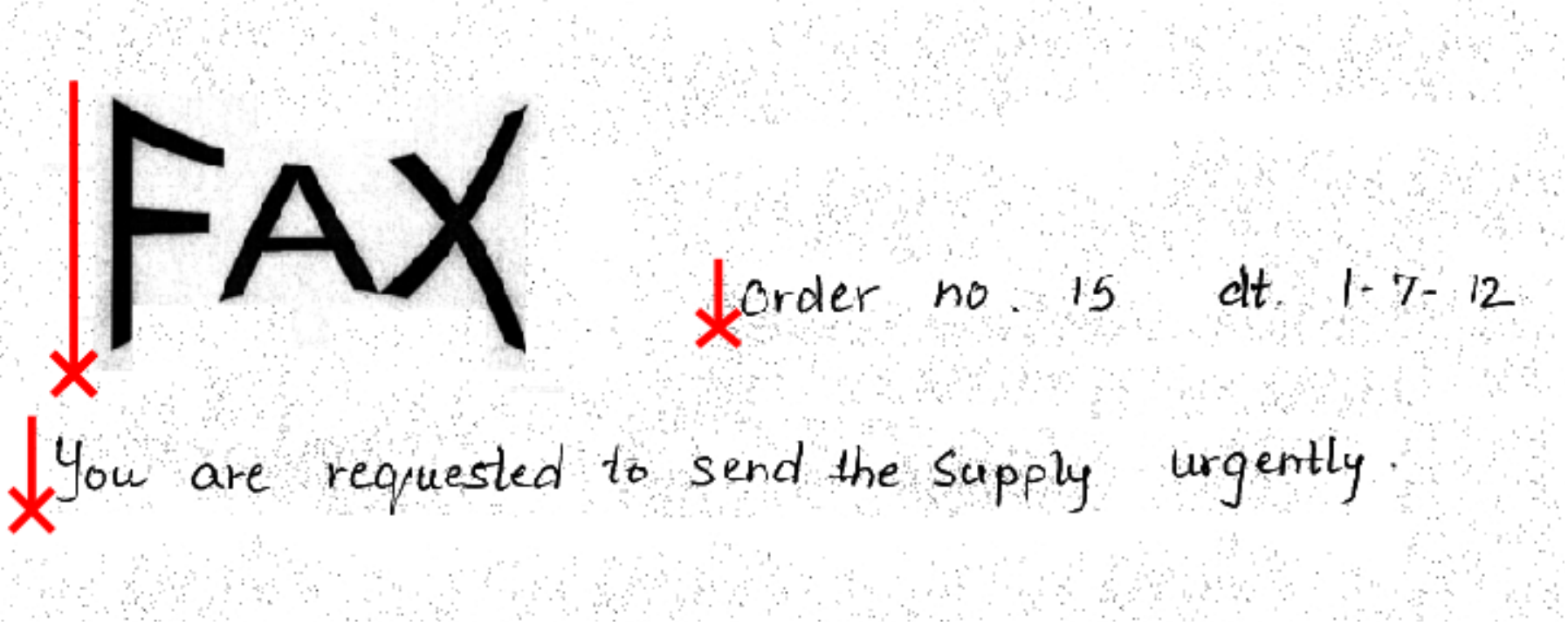} \\
  \small{(b)} \includegraphics[width=2.5in]{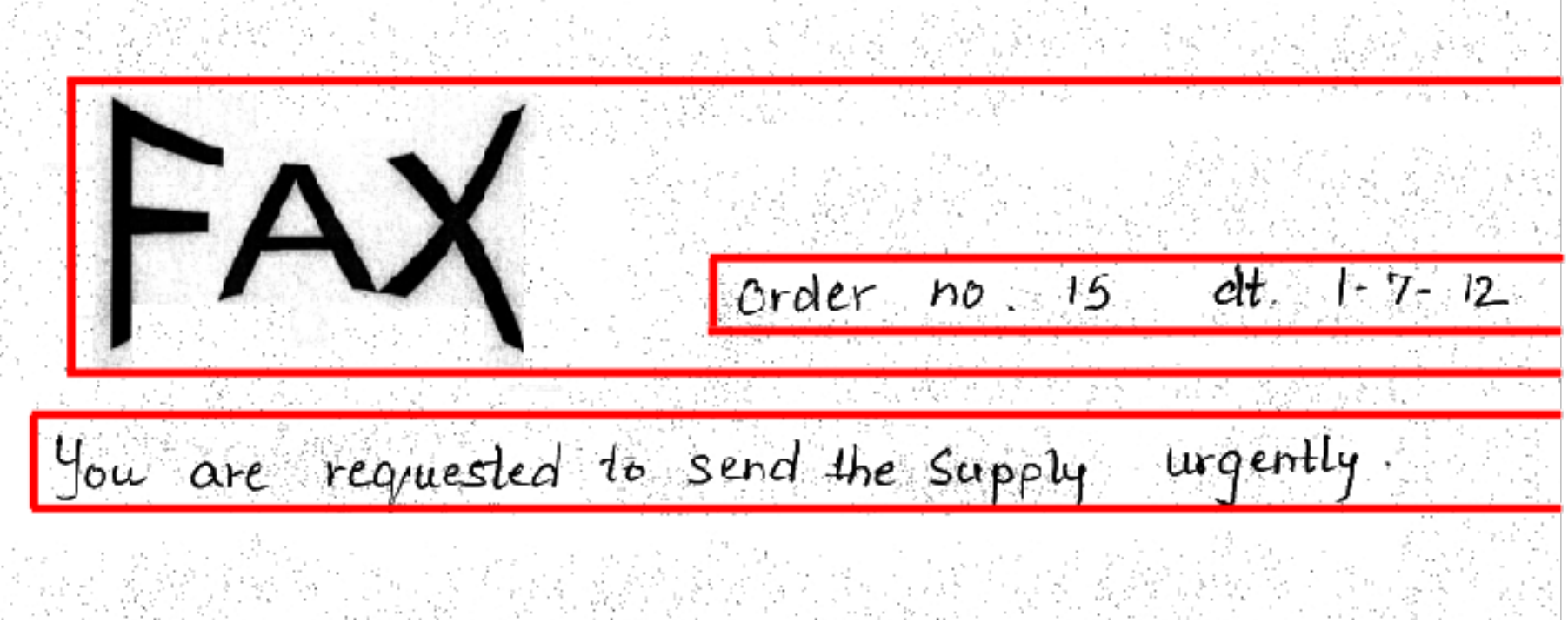} \\
  \small{(c)} \includegraphics[width=2.5in]{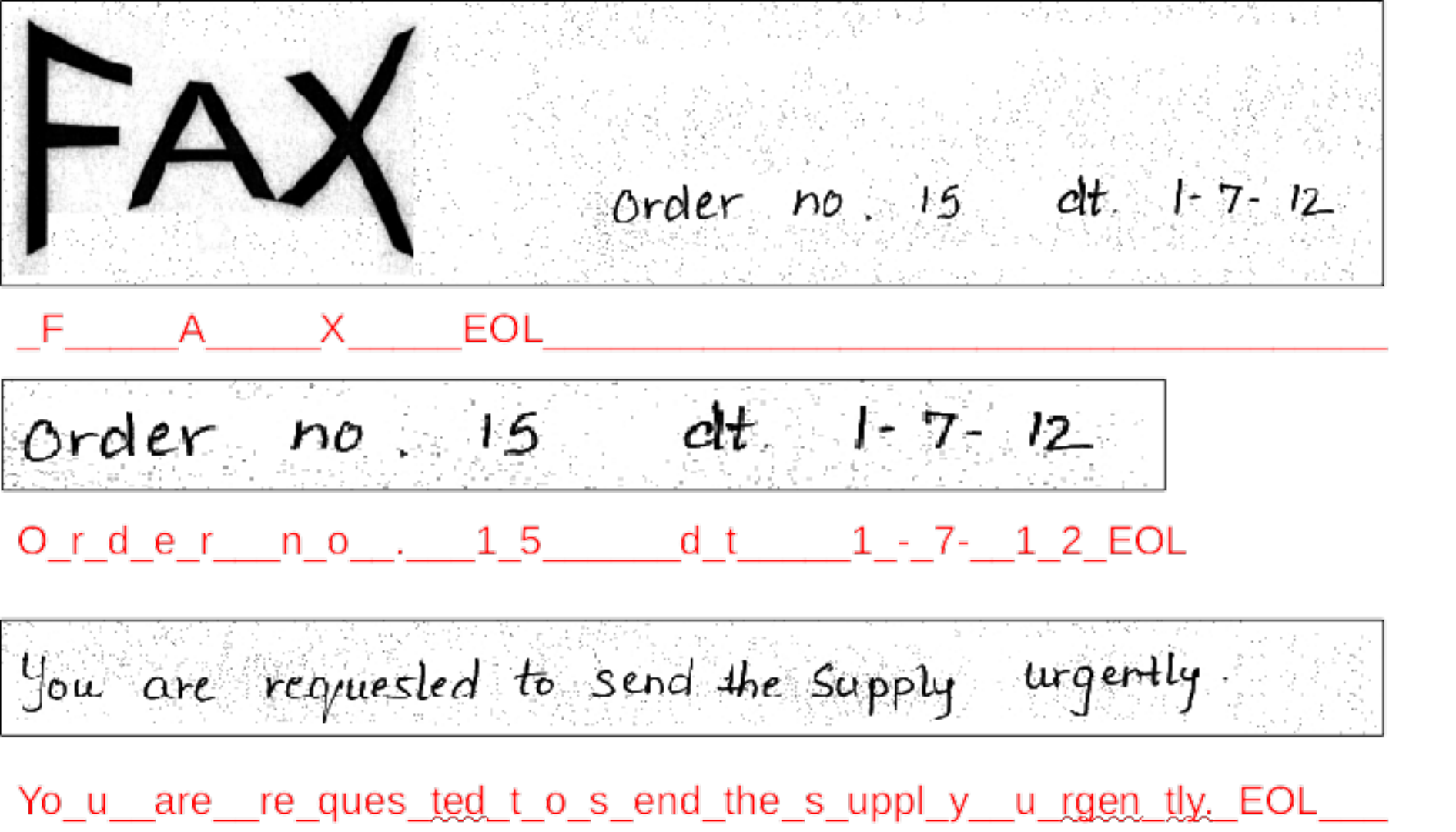}
  \caption{Description of the triplet-based localization techniques. (a) Detection of left-side triplet objects. (b) Extraction of corresponding text lines. (c) Recognition of text lines with End-of-line characters (EOL).}
  \label{fig:triplets}
\end{figure}

\subsection{Line detection with left-side triplets}
\label{sec:triplets}
The first step detects the left-side of each lines through the network described in Section \ref{sec:objectLocalization}. The model predicts three position values, i.e. $K{=}3$: 2 coordinates for the lower left corner plus the text height. Additionally, a prediction confidence score is output.

We also compare this method with two competing strategies: 

i) point localization \cite{moysset2016points}, $K{=}2$, where only the x and y coordinates of lower left points are detected, 

ii) and full box localization \cite{moysset2016learning}, where $K{=}4$ and the x and y coordinates of the bottom-left corners of the text line bounding boxes are predicted with the width and the height of the text lines. 

We also found that expanding the text box by a 10 pixel margin 
improves the full-page text recognition rates. 


\subsection{End-of-line detection integrated with recognition}
\label{sec:recognizer}
Detecting only the left side of the text lines and extending it toward the right part of the image as illustrated in Figure \ref{fig:triplets} b) means that for documents with complex layouts, some text from other text lines can be present in the image to be recognized.

For this reason, a 2D-LSTM based recognizer similar to the one described in \cite{pham2014dropout} is trained with the Connectionist Temporal Classification \cite{Graves06connectionisttemporal} (CTC) alignment procedure to recognize the text present in, and only in, the designed text line. We found that the results were slightly improved by adding a End-of-line (EOL) label at the end of the text labels. 
This  means that the network will learn, through the CTC training, to align the text labels with the frames corresponding to these image characters, as usual. But it will also learn to predict when the line is over, mark it with the EOL label, and learn  not to predict anything else on the right side of the image. The context conveyed by the LSTM recurrent layers is responsible for this learning ability.

Two different recognition networks are trained, respectively for French and English. They are trained to recognized both printed and handwritten simultaneously. 

\section{Experimental Setup}
\label{sec:experimental}
\subsection{Datasets}
We evaluate our method on the Maurdor dataset \cite{Brunessaux2014}, which is composed of 8773 heterogeneous documents in French, English or Arabic, both printed and handwritten (Train: 6592, Validation: 1110, Evaluation: 1071). Because the annotation is given at paragraph level, we used the technique described in \cite{bluche2014automatic} to check the quality of line candidates with a constrained text recognition in order to obtain annotation at line level. All these lines are used to train the text recognizers described in section \ref{sec:recognizer} and, on the test set, for the recognition experiments in Table \ref{tab:reco}.

For training the text line detection systems, only the 5308 pages for which we are confident enough that all the lines are detected are kept (Train: 3995, Validation: 697, Evaluation: 616). This subset is also used for the precision experiments shown in Tables \ref{tab:precisenessPoint} and \ref{tab:precisenessTriplet}.

Finally, for the end-to-end evaluation results shown in Table \ref{tab:fullPage}, we kept all the original documents of the test set in only one language, in order to avoid the language identification problem. We obtain 507 pages in French and 265 pages in English.

\subsection{Metrics}
Three metrics were used for evaluation : 

\begin{enumerate}
	\item \textbf{F-Measure} metrics is used in Tables \ref{tab:precisenessPoint} and \ref{tab:precisenessTriplet} in order to measure precision the  of detected objects being in the neighbourhood of reference objects. A detected object $l$ is considered as correct if it is the closest hypothesis from the reference object $t$ and if $||l^k - t^k|| < T$ for all $k \in [0,K]$ where K is the number of coordinates per object set to 2 for Table \ref{tab:precisenessPoint} (points) and to 3 in Table \ref{tab:precisenessTriplet} (triplets) and $T$ is the size of the acceptance zone given as a proportion of the page width. 
	\item \textbf{Word Error Rate} (WER) metrics is the word level Levenshtein distance \cite{levenshtein1966binary} between recognized and reference sequences of text.
	\item  \textbf{Bag of Word} (BOW) metrics is given at page level as a F-Measure of words recognized or not in the page. As explained in \cite{pletschacher2015europeana}, it is a proper metric to compute recognition rate at page level because it does not need any alignment or ordering of the text lines that can be ambiguous for unconstrained documents.
	
\end{enumerate}

\subsection{Hyper-parameters}
We trained with the RmsProp optimizer \cite{tieleman2012lecture} with an initial learning rate of $10^{-3}$ and dropout after each convolutional layer. The $\alpha$ parameter is set to $\alpha{=}1000$ for matching (solving for X) and to $\alpha{=}100$ for gradient computation.

\section{Experimental results}
\label{sec:results}
\subsection{Precision of the object localizations}
Similarly to what is described in \cite{moysset2016points}, we observed some instability in the position of the predicted objects when trying to detect boxes. Our intuition is that precisely detecting objects which ends outside of the convolutional receptive field of the outputs is difficult. 

Characters may have a size of 1 or 2 mm in standard printed pages, corresponding to 0.005 and 0.01 as a proportion of the page width. Interlines may have similar sizes. Therefore, it is important that the position prediction is close enough in order not to harm the text recognition process.

The method described in \cite{moysset2016points} was dealing with this problem by detecting separately the bottom-left and top-right points and posteriorly pairing them. We observed that the precision was not harmed by the detection of triplets of coordinates (left, top, bottom). 

In Table \ref{tab:precisenessPoint}, we show the F-measure of the detection of left-bottom points for several acceptance zone sizes. The results emphasize that detecting full text boxes reduces precision. Meanwhile, the precision of bottom-left point prediction is equivalent when the network is trained to detect triplets and not points.

Table \ref{tab:precisenessTriplet} shows the same experiment with a 3D acceptance zone defined on the triplet positions, showing the same improved results for the triplet detection for small acceptance zones.

\begin{table}
\begin{center}
\caption{Comparison of the F-Measure scores for the detection of bottom-left points with respect to the acceptance zone size for networks trained to detect points, triplets or boxes. Acceptance zones are given as proportion of page width.}
\label{tab:precisenessPoint}
\begin{tabular}{lccccc}
Network & 0.003 & 0.01  & 0.03 &  0.1  \\ 
\arrayrulecolor{cwblue1} \toprule
Box network (\cite{moysset2016learning}, $K{=}4)$ & 6.8\% & 45.0\% & 82.8\% & 89.9\% \\ 
Point network (\cite{moysset2016points}, $K{=}2$) & 10.7\% & 57.4\% & 85.7\% & 91.7\% \\ 
Triplet network (Ours, $K{=}3$) & 11.2\% & 58.4\% & 87.0\% & 92.6\% \\ 
\end{tabular}
\end{center}
\end{table}

\begin{table}
\begin{center}
\caption{Comparison of the F-Measure scores for the detection of left-side triplets with respect to the acceptance zone size for networks trained to detect triplets or boxes. Acceptance zones are given as proportion of page width.}
\label{tab:precisenessTriplet}
\begin{tabular}{lccccc}
Network & 0.003 & 0.01  & 0.03 &  0.1  \\ 
\arrayrulecolor{cwblue1} \toprule
Box network (\cite{moysset2016learning}, $K{=}4$)  & 3.4\% & 24.6\% & 71.4\% & 89.7\% \\ 
Triplet network (Ours, $K{=}3$) & 4.2\% & 47.2\% & 84.8\% & 92.4\% \\ 
\end{tabular}
\end{center}
\end{table}

\subsection{Detection of the line end with the text recognizer}
In Table \ref{tab:reco} we compared two text line recognizers trained respectively on the reference text line images and on text line images defined only by the left sides coordinates of the text line and extended toward the right end of the page. These two recognizers are evaluated in both cases with the WER metric.

While the network trained on reference boxes is obviously not working well on extended test images, we see that the network trained on extended lines works on both tasks nearly as well as the network trained on reference boxes.This confirms that we can rely on the text recognizer to ignore the part of the line that does not belong to the text line.

\begin{table}
\begin{center}
\caption{Text recognition Word Error Rates (WER) for networks trained/evaluated on reference boxes or box defined only by their left sides. }
\label{tab:reco}
\begin{tabular}{c|cc}
 & Evaluated on & Evaluated with \\
 & reference boxes & left-sides only \\
\arrayrulecolor{cwblue1} \toprule
Trained on reference boxes & 9.0\% & 46.7\% \\
Trained with left-sides only & 10.6\% & 9.8\% \\
\end{tabular}
\end{center}
\end{table}

\subsection{Full page text recognition}
Finally, we compared our method with baselines and concurrent approaches for  full page recognition. The evaluation was carried out using the BOW metric and is shown on Table \ref{tab:fullPage}.
We show that the proposed methods yield good results on both the French and English subsets, consistently overpassing the document analysis baselines based on image processing, the object localisation baseline and the concurrent box detection and paired point detection systems.
Some illustrations of the left-side triplets detection alongside with the final full-page text recognition are given in Figure 3 and emphasize the ability of the system to give good results on various types of documents.

\begin{table}
\begin{center}
\caption{Comparison of full-page recognition systems with the Bag of Words (BOW) metric on the French and English documents of the Maurdor dataset. }
\label{tab:fullPage}
\begin{tabular}{c|cc}
System & French dataset & English dataset \\
\arrayrulecolor{cwblue1} \toprule
Shi et al. \cite{Shi2009a} & 48.6\% & 30.4\% \\
Nicolaou et al. \cite{Nicolaou2009} & 65.3 \% & 50.0 \% \\
\arrayrulecolor{cwblue1} \toprule
Erhan et al. & 65.3 \% & 50.0 \% \\
Multibox \cite{erhan2014scalable} & 27.2\% & 14.8\% \\
Multibox \cite{erhan2014scalable} (optimized) & 32.4\% & 36.2\% \\
\arrayrulecolor{cwblue1} \toprule
Box network \cite{moysset2016learning}  & 71.2\% & 71.1\% \\
Points network \cite{moysset2016points} & 71.7\% & 72.3\% \\
\arrayrulecolor{cwblue1} \toprule
Triplet network (proposed) & 79.9\% & 79.1\% \\
\end{tabular}
\end{center}
\end{table}

\section{Conclusion}
We described a full page recognition system for heterogeneous unconstrained documents that is able to detect and recognize text in different languages. The use of a neural network localisation process helps to be robust to the intra-dataset variations. In order to simplify the process and to gain both in precision and in preciseness, we focus on predicting the starting point (left) of the text line bounding boxes and leave the prediction of the end point (right)  to a 2D-LSTM based text recognizer. We report excellent results on the Maurdor dataset and show that our method outperform both image-based and concurrent learning-based methods.

\clearpage

\begin{figure*}[th!] \centering
  \label{fig:IlusResultsFullPage} 
  \caption{Illustration of full-page recognition results.}
  \subfloat[Input+Localization results]{ 
    \includegraphics[width=7cm]{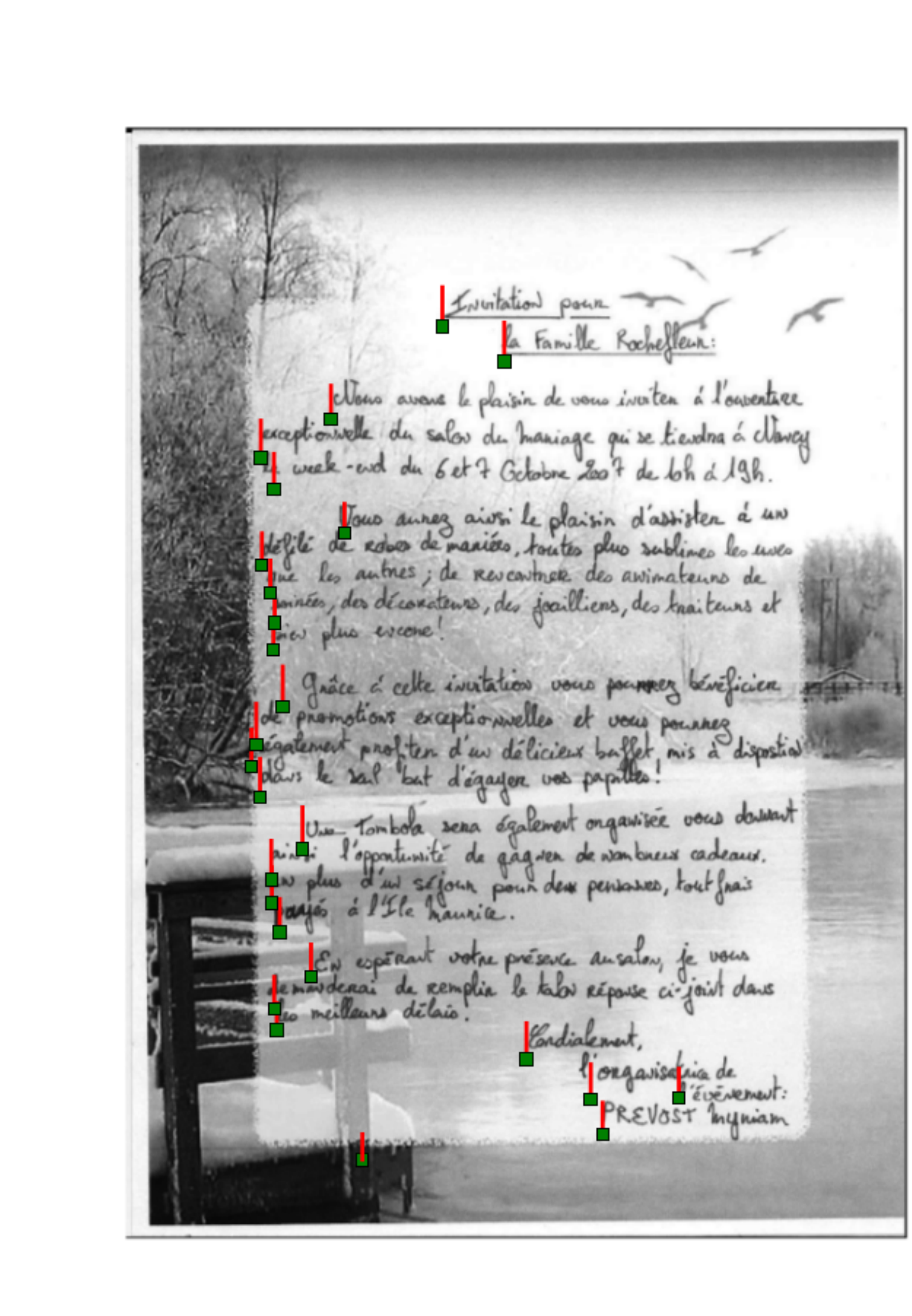}
  }
  \subfloat[Input+Localization results]{
    \includegraphics[width=5cm]{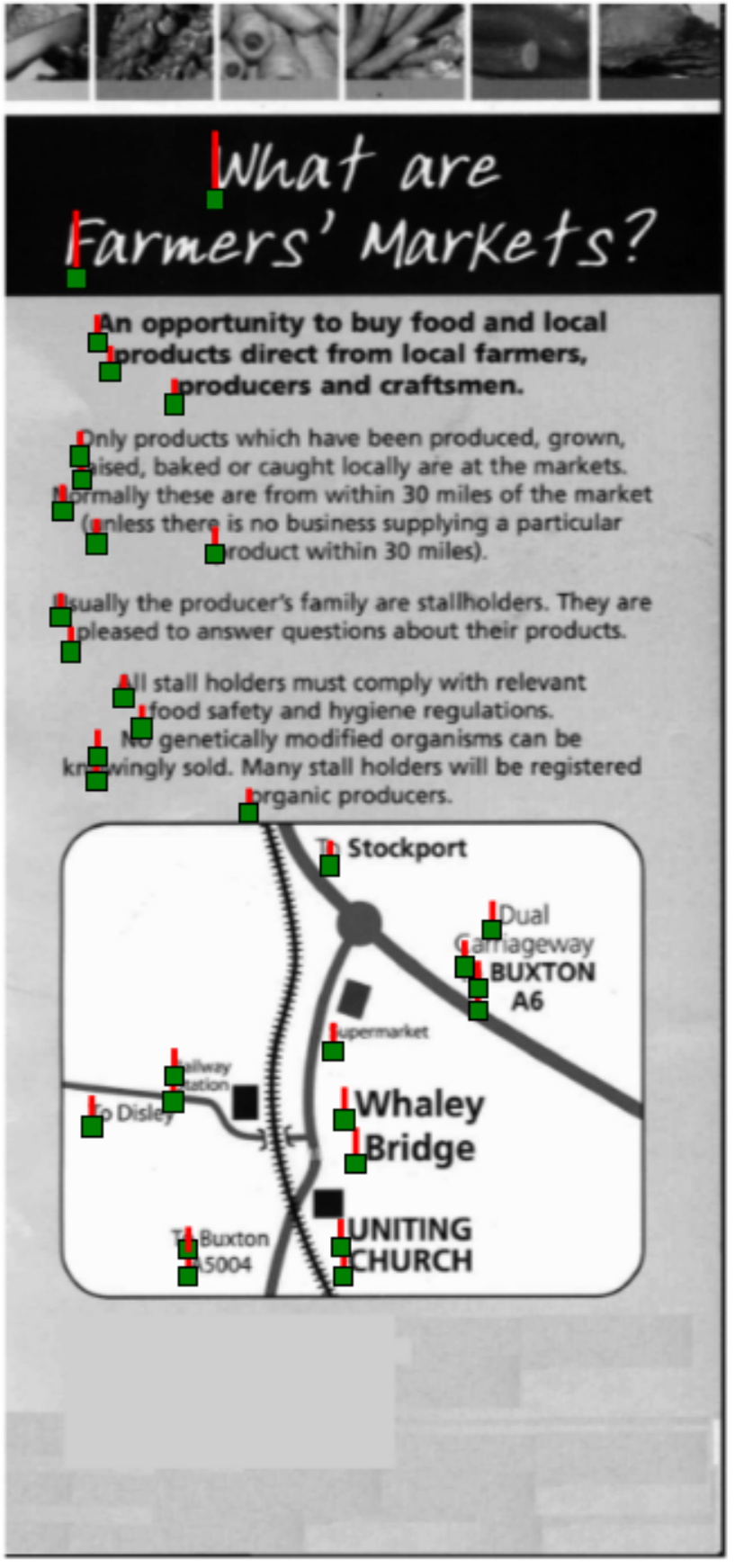}
  }
  \\
  \subfloat[Recognized text]{
   \label{fig:imgIlus1}
  \begin{minipage}[b]{8cm}
    \begin{small}
    \begin{itemize}
      \item Invitation pour mon∞
      \item la Famille Rochefleur :∞
      \item Nous avons le plaisir de vous inviter \`{a} l'ouverture∞
      \item  exceptionnelle du salon du mariage qui se tiendra \`{a} chargey∞
      \item le verte-end du 6 et 7 Octobre 2007 de 20h \`{a} 19h.∞
      \item Vous aurez ainsi le plaisir d'assister \`{a} un∞
      \item d\'{e}fil\'{e} de robes de mari\'{e}es, toutes plus oublimes les unes.∞
      \item que les autres ; de rencontrer des animateurs de∞
      \item sin\'{e}es, des d\'{e}coracteurs, des joailliers, des traiteuns et :∞
      \item bien plus encore !∞
      \item Gr\^{a}ce \`{a} cette invitation vous pourrez b\'{e}n\'{e}ficier de :∞
      \item de prenmations exceptionnelles et vous pourrez∞
      \item \'{e}galement profiter d'un d\'{e}licieux buffet, mis \`{a} disposition!∞
      \item dans le sait bout d'\'{e}galer vos papilles!∞
      \item Une Tombola sera \'{e}galement organis\'{e}e vous donnant.∞
      \item ainsi l'opportunit\'{e} de gagner de nombreux cadeaux.∞
      \item En plus d'un s\'{e}jour pour deux personnes, tout frais∞
      \item pass\'{e}s \`{a} l'Ile haunit\'{e}.∞
      \item En esp\'{e}rant votre pr\'{e}sence au salon, je vous∞
      \item demanderai de remplir le talon r\'{e}ponse ci-joint dans∞
      \item les meilleurs d\'{e}lais !∞
      \item Cordialement,∞
      \item l'organisatrice de∞
      \item LEU******∞
      \item PREVOST Myriam∞
      \item *******∞
    \end{itemize}
    \end{small}
  \end{minipage}
  }
  \subfloat[Recognized text]{   
  \label{fig:imgIlus3}
  \begin{minipage}[b]{8cm}
    \centering
    \begin{small}
    \begin{itemize}
      \item DY∞
      \item ****4481991991∞
      \item An opportunity to buy food and local∞
      \item products direct from local farmers,∞
      \item 1cts direct from local farmers,∞
      \item Only products which have been produced, grown,∞
      \item raised, baked or caught locally are at the markets.∞
      \item formally these are from within 30 miles of the market∞
      \item (unless there is no business supplying a particular∞
      \item product within 30 miles).∞
      \item Usually the producer's family are stallholders. They are∞
      \item pleased to answer questions about their products.∞
      \item All staall holders must comply with relevant∞
      \item food safety and hygiene regulations.∞
      \item No genetically modified organisms can be∞
      \item knowingly sold. Many staall holders will be registered∞
      \item organic producers.∞
      \item To Stockport∞
      \item Dual∞
      \item Carriageway∞
      \item to BUXTON∞
      \item A6∞
      \item supermarket∞
      \item Mailway∞
      \item ∞
      \item To Disien-∞
      \item Whalay∞
      \item Bridge∞
      \item To Burton∞
      \item A5004∞
      \item UNITTING∞
      \item CHURCH∞
    \end{itemize}
    \end{small}
  \end{minipage}
  }
\end{figure*}

\clearpage

\begin{figure}[th!]
  \centering{\footnotesize{(e) Input+Localization results}}
  \\
  \includegraphics[width=6cm]{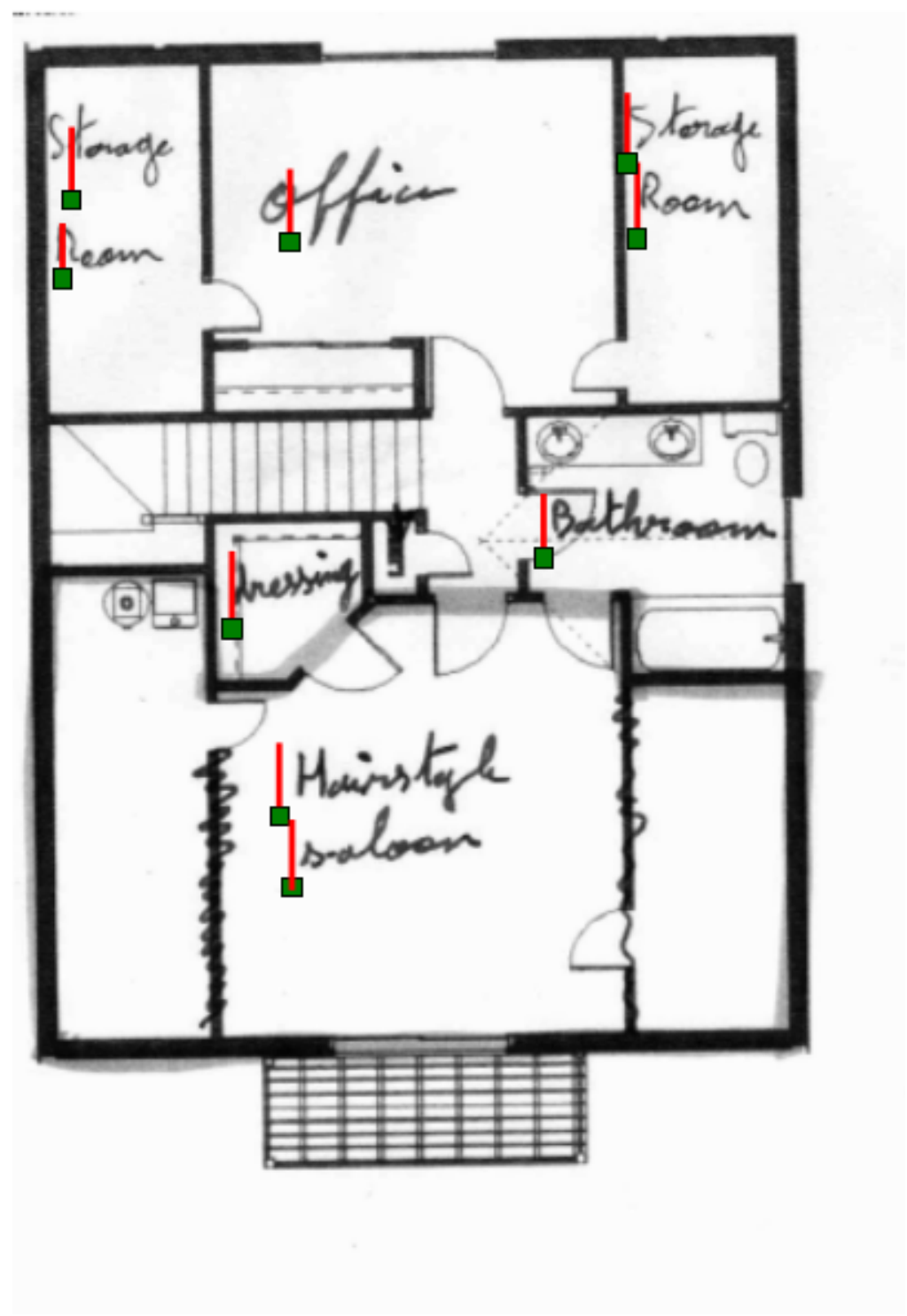}
  \\
   \centering{\footnotesize{(f) Recognized text}}
   \label{fig:imgIlus1}
    \begin{small}
    \begin{itemize}
       \item tonage∞
       \item Room∞
       \item offi***∞
       \item Storage∞
       \item Room∞
       \item Faithroom∞
       \item Merrill∞
       \item Mainstake∞
       \item Saloon∞
    \end{itemize}
    \end{small}
\end{figure}





%

\bibliographystyle{splncs03}
\bibliography{tripletNN}

\end{document}